\documentclass[conference]{IEEEtran}

\usepackage{amssymb,amsmath}
\usepackage{multirow}
\usepackage[utf8]{inputenc}
\usepackage[vietnam,english]{babel}
\usepackage{tabularx}

\usepackage{cite}
\usepackage[english,vietnam]{babel}

\ifCLASSINFOpdf
 \usepackage[pdftex]{graphicx}
\else
\fi
\begin{document}
\selectlanguage{english}
%
\title{Vietnamese Transition-based Dependency Parsing with Supertag Features}

\author{\IEEEauthorblockN{Kiet V. Nguyen}
\IEEEauthorblockA{Department of Information Science and  Engineering\\
University of Information Technology\\
Vietnam National University - Ho Chi Minh City, Vietnam\\
Email: kietnv@uit.edu.vn}
\and
\IEEEauthorblockN{Ngan Luu-Thuy Nguyen}
\IEEEauthorblockA{Faculty of Computer Science\\
University of Information Technology\\
Vietnam National University - Ho Chi Minh City, Vietnam\\
Email: ngannlt@uit.edu.vn}}


%


\maketitle

\begin{abstract}
In recent years, dependency parsing is a fascinating research topic and has a lot of applications in natural language processing. In this paper, we present an effective approach to improve dependency parsing by utilizing supertag features. We performed experiments with the transition-based dependency parsing approach because it can take advantage of rich features. Empirical evaluation on Vietnamese Dependency Treebank showed that, we achieved an improvement of 18.92\% in labeled attachment score with gold supertags and an improvement of 3.57\% with automatic supertags.  \\
\end{abstract}


%
\IEEEpeerreviewmaketitle

\section{Introduction}
Dependency parsing is one of the basic and important natural language processing problems. Outputs of dependency parsing are dependency structures which are relations of two words. In recent years, dependency parsing has a lot of real world applications such as Question Answering \cite{QA_Application}, Machine Translation \cite{MT_Application}, Information Extraction \cite{IE_Application}, Opinion Mining \cite{SA_Application} and so on. There are a lot of research related to dependency parsing. Large and prestigious conferences in the field of computational linguistics, including ACL, EACL, and COLING, have constantly provided many tutorials on dependency parsing \cite{Tutorial_2006, Tutorial_2010, Tutorial_2014_2, Tutorial_2014_1, Tutorial_2013}. Especially, the 2006 and 2007 CoNLL Shared Tasks \cite{CoNLL_2006, CoNLL_2007}  attracted a lot of research works in study on data-driven dependency parsing on many languages.\\ 

\begin{figure}[h]
\centering
\includegraphics[scale=0.5]{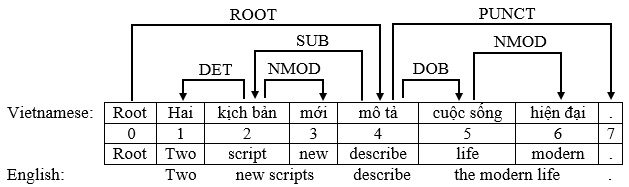}
\caption{A dependency graph example for a Vietnamese sentence which is borrowed from \cite{DP_Vi_4}.}
\label{fig:sentence_1}
\end{figure}

\selectlanguage{vietnam}
The problem of dependency parsing is described as: Given an input sentence with the length of n words S = $w_0$, $ w_1$, $w_2$, ..., $w_n$, where $w_0$ = ROOT, the goal of dependency parsing is to analyze this sentence into a labeled dependency graph as described in Figure \ref{fig:sentence_1}. A triple (h, d, l) represents a labeled arc, where node h is the head (or parent) of the modifier (or child, dependent) d, and their syntactic dependency relation is l. For instance, the label SUB describes a subject dependency relation between the head word \textit{mô tả/describes} and the modifier \textit{kịch bản/scripts}. It is important to note that each word must have exactly one incoming arc and the zero node indicates the root of the sentence.\\

\selectlanguage{english}

A wide range of research works on dependency parsing has demonstrated the effectiveness of utilizing rich feature representations, for examples, fined-grained POS features \cite{Fined_Grained_Features}, rich non-local features \cite{Rich_Non_local_Features}, supertag features \cite{Supertag_Features}. But state-of-the-art parsing systems for the Vietnamese language generally use only two basic kinds of features: word form and part-of-speech (POS) tag. In this paper, we can take advantage of supertag features to improve Vietnamese dependency parsing. Supertags are complex features that capture fined-grained syntactic phenomenon and long-distance dependencies \cite{Supertag_Features}.\\

The contributions of this paper to the development of dependency parsing for Vietnamese are summarized as follows.\\

\begin{itemize}
\item First, we demonstrate the effectiveness of supertags as a new kind of feature for Vietnamese dependency parsing;\\
\item Second, we propose three types of design for supertags which are suitable for Vietnamese transition-based dependency parsing;\\
\item Third, we has built a Vietnamese dependency parsing tool which is trained on Vietnamese Dependency Treebank (VnDT) \ \cite{DP_Vi_2}. This system achieves the highest accuracy and becomes a state-of-the-art dependency parser.\\
\end{itemize}


The rest of this paper is oganized as follows. Section 2 discusses the related works on transition-based dependency parsing, Vietnamese dependency parsing, and parsing features. Section 3 describes three designs of supertags for improving Vietnamese dependency parsing. Section 4  shows the experimental parsing results reported using the cross-validation scheme. 
Finally, we draw a conclusion and future directions for dependency parsing for Vietnamese language in Section 5. 

\section{Related Works}
\subsection{Transition-based Dependency Parsing}
\begin{figure}[!h]
\selectlanguage{english}
\centering
\includegraphics[scale=0.55]{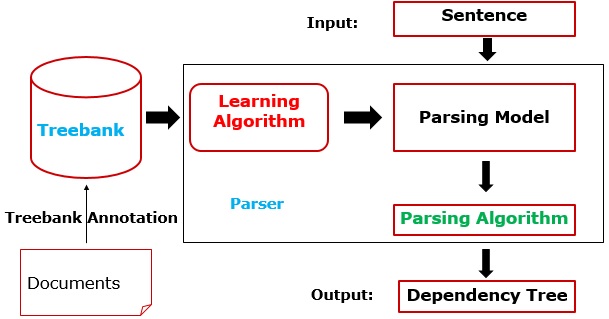}
\caption{An architecture of data-driven dependency parsing system.}
\label{fig:dpmodel}
\end{figure}

Dependency parsing has two main approaches: grammar-driven and data-driven. Research works on dependency parsing mainly concentrate on data-driven models for its historical development. An architecture of data-driven model for dependency parsing is shown in Figure \ref{fig:dpmodel}. There are three important modules including the learning algorithm, the parsing model, and the parsing algorithm. Depending on the parsing algorithm, data-driven dependency parsing is divided two types including graph-based and transition-based. In this paper, we used transition-based dependency parsing to build the Vietnamese dependency parsing system with new features for the learning module.\\ 

Transition-based dependency parsing systems \cite{Nirve_2006}, \cite{Nirve_2006_1}, \cite{Nirve_2006_2}, \cite{Nivre_2007}, are among state-of-the-art parsing systems because of high accuracy and speed in many natural languages. One of the well-known data-driven dependency parsings is MaltParser, an open-source system \cite{Nirve_2006} that has a lot of parameters for customization. It has nine transition-based parsing algorithms which are divided into three parsing algorithm groups: Nirve's algorithm, Convington's algorithms and Stack algorithms \cite{MaltOptimizer} with a wide range of complex feature models.  Maltparser used two libraries for machine learning: LIBSVM \cite{LIBSVM} and LIBLINEAR \cite{LIBLINEAR}. MaltParser can be trained to parse any language as long as we have a treebank in CoNLL format \cite{CoNLL_2006, CoNLL_2007}. In this paper, we employ new features - supertags to improve the parsing for the Vietnamese language.\\

MaltOptimizer \cite{MaltOptimizer} is a automatic tool developed to optimize components of parsers developed using MaltParser. MaltOptimizer can find optimal parameters of each component, it includes an analysis of the dependency treebank and three optimal phases. When performing MaltParser for a new language or domain, there are essentially three components of the optimal system that need to be found: parsing algorithm, feature model, and learning algorithm. In our experiment, we used MaltOptimizer to find suitable aspects for Vietnamese dependency parsing.\\

\subsection{Vietnamese Dependency Parsing}

On Vietnamese, there are few research works on dependency parsing. Some research works \cite{DP_Vi_1, DP_Vi_2} developed Vietnamese dependency treebank from phrase-structure treebank VietTreebank. Thi et al. \cite{DP_Vi_1} achieved 73.03\%
and 66.35\% in the unlabeled and labeled attachment scores given by MaltParser using the gold standard POS tags. Nguyen et al. \cite{DP_Vi_2} achieved an unlabeled attachment score of 79.08\% and a labeled attachment score of 71.66\%. All accuracies of these parsers are under 80\% in unlabeled attachment score. There are still many challenges for building Vietnamese dependency parsers with higher performance.\\

According to the error analysis results for Vietnamese dependency parsing in \cite{DP_Vi_4}, dependency parsing tends to have lower accuracies on long  dependencies. We can see that a lot of dependency relations (noun modifier, verb modifier, subject, direct object, root, adjective modifier, coordination, conjunction, indirect object) have lower accuracies than others, as shown in Table \ref{Relation_Accuracy} (the figures in this table are from the paper \cite{DP_Vi_4}). Therefore, we focus on finding new features to solve above problems related to these dependency relations.\\

\begin{table}[!h]
\centering
\caption{PRECISION/RECALL FOR DEPENDENCY RELATIONS. DR = DEPENDENCY RELATION; DLA = DEPENDENCY LENGTH AVERAGE.}
\label{Relation_Accuracy}
\begin{tabular}{|c|c|c|c|}
\hline
\textbf{DR} & \textbf{DLA} & \textbf{Precision} & \textbf{Recall} \\ \hline
NMOD        & 1.83         & 79.04              & 75.09           \\ \hline
VMOD        & 2.58         & 60.70              & 58.22           \\ \hline
SUB         & 3.57         & 65.70              & 67.29           \\ \hline
DOB         & 1.63         & 68.76              & 64.00           \\ \hline
ROOT        & 5.62         & 79.92              & 79.84           \\ \hline
AMOD        & 1.50         & 72.01              & 69.12           \\ \hline
COORD       & 5.64         & 50.94              & 50.74           \\ \hline
CONJ        & 2.43         & 70.32              & 69.85           \\ \hline
IOB         & 2.80         & 27.45              & 37.95           \\ \hline
\end{tabular}
\end{table}

Most research works on Vietnamese dependency parsing have yet taken advantage of linguistic knowledge for the design of dependency parsing features. In this paper, we proposed and designed new features to improve our dependency parsing for Vietnamese in section 3.\\

\subsection{Corpus: Vietnamese Dependency Treebank}

As shown in Figure \ref{fig:dpmodel}, in order to build a data-driven parsing system, we must have a large set of dependency trees - treebank. Vietnamese Dependency Treebank (VnDT) is a treebank containing dependency structures converted from Vietnamese Treebank \cite{VietTreebank} following the approach of Nguyen et al. \cite{DP_Vi_2}. There are 33 dependency relations in the VnDT treebank.  The percentage of non-projective structures in VnDT is 4.49\%. The percentage of sentences with length less than 30 words is 80\%. The proportion of sentences with length over 20 words accounts for 45.61\%. In this paper, we conducted experiments on this treebank. \\

\subsection{Transition-based Dependency Parsing with Supertag Features}
A wide range of research works \cite{Yamada_2003, Nivre_2007, Sagae_2010, Goldberg_2010} have shown that  transition-based dependency parsing utilizes rich feature representations. Most experiments on dependency parsers employed two types of basic features: word forms and part-of-speech tags. We should find a different feature type which can improve the accuracy of dependency parsing systems on a new language or domain.\\

Supertags are lexical templates imposed complex constraints in a local context. Supertags are extracted from dependency structure treebank and encode rich syntactic information. There are many parsing systems improved with supertags, for instance, the English dependency parsing \cite{Supertag_Features}, the German dependency parser with Weighted Constraint Dependency Grammar (WCDG) \cite{Foth}, Hindi transition-based dependency parsing with supertags based on CCG lexicon \cite{Ambati}. For English, Ouchi et al. \cite{Supertag_Features} proposed two supertag models for transition-based dependency parsing with an improvement of 1.3\% in unlabeled attachment score. In our experiments, we used supertags to improve our dependency parsing. We present in detail our supertag designs in the next section.

\section{Supertag Designs for Vietnamese}

In this paper, we proposed utilizing linguistic features to improving Vietnamese dependency parsing. Supertags are labels for tokens much like part-of-speech tags but they also encode fine-grained information of syntax. The major difficulty of the supertag designs is to find a tagset that balance between granularity and predictability. We would like to increase the number of supertags considerably and make it more difficult to predict automatically. We designed supertags based on the two models proposed by Ouchi et al. \cite{Supertag_Features} and error analysis for Vietnamese dependency parsing by Nguyen et al. \cite{DP_Vi_4}. To suit Vietnamese dependency parsing, we design supertags related to dependency relations such as NMOD, VMOD, SUB, DOB, ROOT, AMOD, COORD, CONJ, and IOB. Because these dependency relations have low accuracies as analyzed in the paper \cite{DP_Vi_4}. We describe how to create three supertag models in the following.

\begin{table}[]
\centering
\caption{Design of Three Models of The Supertag for the sentence}
\label{my-label}
\begin{tabular}{|c|c|c|c|}
\hline

\textbf{Word}                                                 & \textbf{Model 0} & \textbf{Model 1} & \textbf{Model 2} \\ \hline

\begin{tabular}[c]{@{}c@{}}\selectlanguage{vietnam}(Hai)\\ (Two)\end{tabular}           & DET                 & DET                 & DET                 \\ \hline
\begin{tabular}[c]{@{}c@{}}\selectlanguage{vietnam}(kịch bản)\\ (scripts\}\end{tabular} & SUB/R               & SUB/R+L\_R          & SUB/R+L\_R          \\ \hline
\begin{tabular}[c]{@{}c@{}}\selectlanguage{vietnam}(mới)\\ (new)\end{tabular}           & NMOD/L              & NMOD/L              & NMOD/L                \\ \hline
\begin{tabular}[c]{@{}c@{}}\selectlanguage{vietnam}(mô tả)\\ (describe)\end{tabular}    & ROOT                & ROOT+L\_R           & ROOT + SUB/L\_DOB/R \\ \hline
\begin{tabular}[c]{@{}c@{}}\selectlanguage{vietnam}(cuộc sống)\\ (life)\end{tabular}    & DOB/L               & DOB/L+R             & DOB/L+R             \\ \hline
\begin{tabular}[c]{@{}c@{}}\selectlanguage{vietnam}(hiện đại)\\ (modern)\end{tabular}   & NMOD/L              & NMOD/L              & NMOD/L              \\ \hline
\textbf{.}                                                    & PUNCT               & PUNCT               & PUNCT               \\ \hline
\end{tabular}
\end{table}


\subsection{Model 0}
In Model 0, we designed a supertag which represents syntactic information as follows: \textit{Dependency\_Label/X}, where X indicates the relative position (direction) of the head of a word, which can be left (L) or right (R). With dependency relations, for examples, NMOD, VMOD, SUB, DOB, ROOT, AMOD, COORD, CONJ, and IOB, the supertags combine with the dependency relation label and the left(L) or right(R) information, else, the supertag is the dependency relation. For example, \textit{\selectlanguage{vietnam}kịch bản/script}’s in the example in Figure 1 has its head in the right direction with a label \textit{SUB}, so its supertag can be represented as \textit{SUB/R}. The number of supertags of Model 0 is 40 tags.\\

\subsection{Model 1}

Model 1 is inherited from Model 0. We also add information about whether a word has any left or right dependents for the dependency labels NMOD, VMOD, SUB, DOB, ROOT, AMOD, COORD, CONJ, and IOB. For instance, the word \textit{\selectlanguage{vietnam}cuộc sống/life} has a right dependent \textit{\selectlanguage{vietnam}hiện đại/modern}, so we encode it as \textit{DOB/L+R}, where the part before \textit{+} specifies the head information (\textit{DOB/L}) and the part afterwards (\textit{L}) specifies the position of the dependent (\textit{L} for left, \textit{R} for right). When a word has its dependents in both left and right directions, such as the word \textit{\selectlanguage{vietnam}mô tả/descibe} in Figure 1, we combine them using an underline, as in: \textit{ROOT+L\_R}. On Vietnamese Dependency Treebank, Model 1 has 47 supertags.\\

\subsection{Model 2}

Model 2 is inherited from Model 1. For verbs, we add  obligatory dependents of the main verb to dependency relation labels. The obligatory dependents have the following dependency relation labels, \textit{SUB, DOB, PRD and IOB}. For example, \textit{\selectlanguage{vietnam}mô tả/describe} in the example sentence has an obligatory dependent with a label \textit{SUB} in the left direction and \textit{DOB} in the right direction, so its supertag is represented as \textit{ROOT + SUB/L\_DOB/R}. The number of supertags of Model 2 is 63 tags.\\

\begin{figure}[h]
\centering
\includegraphics[scale=0.5]{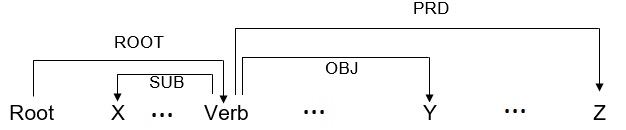}
\caption{Obligatory dependents of the verbs with subjects, objects and predicates}
\label{fig:ObliDependents}
\end{figure}

In this supertag, we focus on the obligatory dependents of the verbs with dependency relations such as subjects, objects (direct and indirect objects), and predicates, as seen in Figure \ref{fig:ObliDependents}. Because it is important to construct dependency graphs and decrease the number of supertags in this model.\\

\selectlanguage{english}

\section{Experiments}
We used Vietnamese Dependency Treebank \cite{DP_Vi_2} to train and test the dependency parsing (MaltParser, version 1.9.0). We conducted evaluations of the parser using the 5-fold cross validation scheme with the average fold size of 2400 sentences. In these experiments, we used automatic supertags, POS tags, and main-verb-attachment POS tags. We used LibSVM for the learning phase. For all experiments, the parameters were set as the followings: c = 0.1 (cost), e = 0.1 (termination criterion), B = 0 (bias). \\
\subsection{Supertagging Experiments}
In this experiment, we used the training and test data extracted from Vietnamese Dependency Treebank according to our proposed supertag templates. We used the training dataset to train the RDRPOSTagger (version 1.2.2) \cite{RDRPOSTagger} to achieve a supertagging model. RDRPOSTagger is a rule-based POS tagging toolkit that automatically constructs transformation rules for POS tagging. It is the current state-of-the-art POSTagger for Vietnamese. 
 
The test dataset was used to calculate the accuracies of the resulted Vietnamese suppertagger.
The experimental results of supertagging are shown in Table \ref{Supertag_Accuracy}.
It can be seen in this table that the supertagging accuracies are around 79-81\% for all three models.  

\selectlanguage{english}
\begin{table}[h]
\centering
\caption{Average accuracies of three supertags models}
\label{Supertag_Accuracy}
\begin{tabular}{|c|c|c|c|}
\hline
\textbf{Fold} & \textbf{\begin{tabular}[c]{@{}c@{}}Model 0\\ (\%)\end{tabular}} & \textbf{\begin{tabular}[c]{@{}c@{}}Model 1\\ (\%)\end{tabular}} & \textbf{\begin{tabular}[c]{@{}c@{}}Model 2\\ (\%)\end{tabular}} \\ \hline
1             & 81.57                                                              & 80.52                                                              & 80.13                                                              \\ \hline
2             & 80.79                                                              & 80.45                                                              & 79.25                                                              \\ \hline
3             & 81.20                                                              & 79.85                                                              & 79.17                                                              \\ \hline
4             & 80.88                                                              & 80.76                                                              & 79.57                                                              \\ \hline
5             & 80.39                                                              & 80.22                                                              & 78.95                                                              \\ \hline
Average       & \textbf{80.97}                                                     & \textbf{80.36}                                                     & \textbf{79.41}                                                     \\ \hline
\end{tabular}
\end{table}

\subsection{Dependency Parsing Experiments}

In this section, we present the experiments on transition-based dependency parsing for Vietnamese. We conducted four dependency parsing experiments including the baseline experiments with basic features (word forms and part-of-speech tags) and the other experiments with the improved parsing models corresponding to three supertag models. All experiments were carried out through 3 steps, including:\\

\begin{itemize}

\item \textbf{Step 1}: We used the MaltOptimizer tool \cite{MaltOptimizer} to find the best parameters:
\subitem + Choosing the parsing algorithm.
\subitem + Creating automatically the feature model from basic feature types (word forms, POS tags, and supertags).
\subitem + Choosing the parameters of the learning algorithm.\\

\item \textbf{Step 2}: We used the transition-based dependency parsing system - MaltParser \cite{MaltOptimizer} to conduct experiments on 5-fold cross validation scheme with settings from MaltOptimizer results in Step 1.\\

\item \textbf{Step 3}: We used the evaluation tool for dependency parsing - MaltEval \cite{MaltEval} to evaluate the Vietnamese dependency parsing on both gold and automatic tags (POS tags and supertags). The experimental results were reported in accuracy measured with two settings: with and without consideration of dependency labels, i.e., UAS (Unlabeled Attachment Score) and LAS (Labeled Attachment Score). UAS is the percentage of tokens that search correctly the heads. LAS is the percentage of tokens for which the parser has predicted correctly both the heads and the dependency labels.\\

\end{itemize}

\begin{table}[h]
\centering
\caption{Accuracy of Vietnamese Dependency Parsing (\%)}
\label{Parsing_Accuraccies}
\begin{tabular}{|l|l|l|l|l|}
\hline
\multicolumn{1}{|c|}{\multirow{2}{*}{\textbf{Model}}} & \multicolumn{2}{c|}{\textbf{Gold Supertag}} & \multicolumn{2}{c|}{\textbf{Automatic Supertag}} \\ \cline{2-5} 
\multicolumn{1}{|c|}{}                                          & UAS                & LAS               & UAS                & LAS               \\ \hline
Baseline                                                        &76.1                    &69.9                   &73.5                    &65.2                   \\ \hline
0                                                    &88.0                    &87.9                   &78.3                    &68.3                   \\ \hline
1                                    &89.9                    &88.8                   &78.5                    &68.8                   \\ \hline
2                                                      & 88.6                   &88.5                   &77.4                    &68.6                   \\ \hline
\end{tabular}
\end{table}

Table \ref{Parsing_Accuraccies} shows the results of Vietnamese transition-based dependency parsing with three supertag features. We begin with baseline features, and add three supertag features (Model 0, Model 1, Model 2). We can see that adding Model 0 features improves the baseline LAS sharply by 18.02\% in the gold supertags while using automatic supertags gives a smaller improvement of 3.07\%. For Model 1, gold supertag features make the bigger contribution to the improvements by 18.92\% than automatic ones 3.57\%. For Model 2, gold supertag features make a bigger improvement (18.6\%) in comparison with the automatic supertag features (3.37\%). We can see increasing results in Table \ref{measure}. With these results, we can see that the three supertag models outperform the baseline in all metrics. In other, the three kinds of supertag features are effective for the Vietnamese transition-based dependency parsing.\\

\begin{table}[h]
\centering
\caption{Difference between supertag and baseline parsing systems}
\label{measure}
\begin{tabular}{|l|l|l|l|l|}
\hline
\multicolumn{1}{|c|}{\multirow{2}{*}{\textbf{Model}}} & \multicolumn{2}{c|}{\textbf{Gold Supertag}} & \multicolumn{2}{c|}{\textbf{Automatic Supertag}} \\ \cline{2-5} 
\multicolumn{1}{|c|}{}                                          & UAS                & LAS               & UAS                & LAS               \\ \hline
0                                                    & +11.92  & +18.02 & +4.78            & +3.07           \\ \hline
1                                    & \textbf{+13.08}            & \textbf{+18.92}          & \textbf{+4.98}   & \textbf{+3.57}  \\ \hline
2                                                       & +12.52  & +18.62 & +3.88            & +3.37           \\ \hline
\end{tabular}
\end{table}

From the calculations in Table \ref{measure}, we can see that differences between parsing results on gold supertags and automatic supertags are quite large. We can see that the parsing accuracy of automatic supertag data may asymptotic the results on gold supertag data. Therefore, the increase in the performance of the supertagger is one of the research work needed to increase the accuracy of the dependency parsing system. On gold supertag data sets, the LAS accuracy is growing faster than UAS and the differences are not significant.\\

In three supertag designs, we can see the supertags of Model 2 encode rich information of syntax. However, Model 1 is more effective than the two other supertag models (Model 0 and 2). Because the number of supertags in Model 2 is larger than others and Model 1 is the model which has a good balance between granularity and predictability. This may be the reason that makes Model 1 the most effective supertag model for Vietnamese dependency parsing.\\ 

\section{Conclusion and Future Directions}
We present an effective approach to improve transition-based dependency parsing using supertag features. We have shown the effectiveness of supertags for Vietnamese transition-based dependency parsing. Our experimental results show that Model 1 is the most effective supertag model for Vietnamese dependency parsing. Futhermore, the supertags have significantly improved the accuracy of Vietnamese dependency parsing with an increase of 18.92\% for gold supertags and 3.57\% for automatic supertags in the LAS score. To our knowledge, this is the state-of-the-art parsing result for the Vietnamese language so far. For gold standard POS tags, we achieved the UAS score of 89.9\% and the LAS score of 88.8\%. On automatically-assigned POS tags and supertags, the scores are 78.5\% and 68.8\% for the UAS and the LAS, respectively.\\

In future research works, we suggest several possible research directions for improvement of the data-driven dependency for Vietnamese:\\

\begin{enumerate}
\item We would like to improve the supertagger with the CRF algorithm \cite{CRF}. With increasing supertagging accuracy, the transition-based dependency parsing will also be improved. We plan to expand the implementation of the supertags by splitting each supertag into subparts. For example, the supertag ROOT+SUB/L\_DOB/R is splitted into ROOT, SUB/L and PRD/R which encode the information of the supertag head, the left dependents, and the right dependents correspondingly.\\

\item We can also implement fine-grained features on POS tags as proposed by Zhou et al \cite{Fined_Grained_Features} to improve Vietnamese dependency parsing. We can create fine-grained features by splitting different POS tags to different levels using hypernym-hyponymy hierarchical semantic knowledge.\\

\end{enumerate}


\end{document}